\RequirePackage{fix-cm}
\documentclass[smallcondensed]{svjour3}     % onecolumn (ditto)
\smartqed  % flush right qed marks, e.g. at end of proof
\usepackage{graphicx}
\usepackage[pagewise]{lineno}

\begin{document}
%\linenumbers
\title{A double competitive strategy based learning automata algorithm%\thanks{Grants or other notes
%about the article that should go on the front page should be
%placed here. General acknowledgments should be placed at the end of the article.}
}
%\subtitle{Do you have a subtitle?\\ If so, write it here}

%\titlerunning{Short form of title}        % if too long for running head

\author{Chong Di$^1$ \and Shenghong Li$^{1*}$ \and Xudie Ren$^1$ \and Yinghua Ma$^1$ \and Bo Zhang$^2$  %    \and
       % Second Author %etc.
}

%\authorrunning{Short form of author list} % if too long for running head

\institute{Chong Di \at
              \email{dichong95@sjtu.edu.cn}           %  \\
%             \emph{Present address:} of F. Author  %  if needed
           \and
           Shenghong Li \at
              \email{shli@sjtu.edu.cn}
            \and
           XuDie Ren \at
              \email{renxudie@sjtu.edu.cn}
           \and
           Yinghua Ma \at
              \email{ma-yinghua@sjtu.edu.cn}
           \and
           Bo Zhang \at
              \email{zhangbo\_bx@sina.com}
           \and
           $^1$ School of Cyber Space Security, Shanghai Jiao Tong University, 800 Dong Chuan Road, Shanghai 200240, China\\
           $^2$ School of Computer Science and Engineering, Nanjing University of Science and Technology, 200 Xiaolingwei Road, Nanjing 210094, China
}
\date{Received: date / Accepted: date}
% The correct dates will be entered by the editor

\maketitle

\begin{abstract}
Learning Automata (LA) are considered as one of the most powerful tools in the field of reinforcement learning. The family of estimator algorithms is proposed to improve the convergence rate of LA and has made great achievements. However, the estimators perform poorly on estimating the reward probabilities of actions in the initial stage of the learning process of LA. In this situation, a lot of rewards would be added to the probabilities of non-optimal actions. Thus, a large number of extra iterations are needed to compensate for these wrong rewards. In order to improve the speed of convergence, we propose a new P-model absorbing learning automaton by utilizing a double competitive strategy which is designed for updating the action probability vector. In this way, the wrong rewards can be corrected instantly. Hence, the proposed Double Competitive Algorithm overcomes the drawbacks of existing estimator algorithms. A refined analysis is presented to show the $\epsilon-optimality$ of the proposed scheme. The extensive experimental results in benchmark environments demonstrate that our proposed learning automata perform more efficiently than the most classic LA $SE_{RI}$ and the current fastest LA $DGCPA^{*}$.
% \PACS{PACS code1 \and PACS code2 \and more}
% \subclass{MSC code1 \and MSC code2 \and more}
\end{abstract}

\section{Introduction}
\label{intro}
Learning automaton(LA) is one of reinforcement approaches. It is a decision maker which acn choose the optimal action and update its strategy through interacting with the random environment \cite{narendra2012learning}. As one of the most powerful tools in adaptive learning system, LA has had a myriad of applications \cite{oommen1996graph}\cite{nicopolitidis2002using}\cite{esnaashari2010data}\cite{wang2014learning}\cite{zhao2015cellular}\cite{jiang2014new}.

\begin{figure}
% Use the relevant command to insert your figure file.
% For example, with the graphicx package use
  \centering\includegraphics{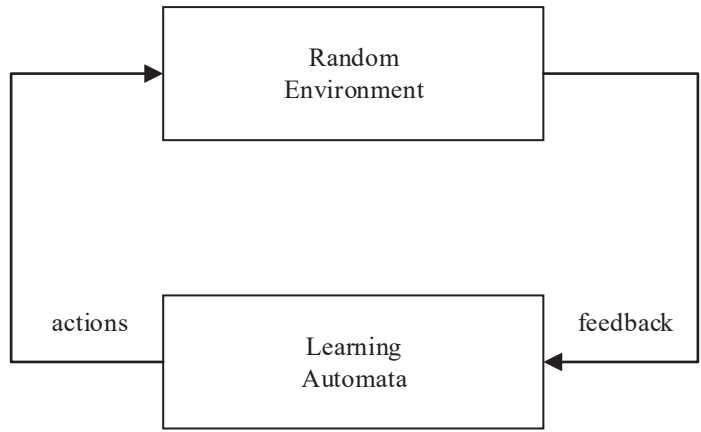}
% figure caption is below the figure
\caption{Learning automata that interact with a random environment \cite{narendra2012learning}}
\label{fig:1}       % Give a unique label
\end{figure}

As illustrated in Fig.\ref{fig:1}, the process of learning is based on a learning loop involving two entities: the random environment(RE) and the LA. In this process, the LA continuously interacts with the RE to get the feedback to its various actions. According to the responses from to the various actions the environment, LA will update the probability vector with a certain method. Finally, the LA attempts to learn the optimal action by interacting with the RE through sufficient iterations.

The first study concerning LA models dates back to the studies by $Tsetlin$ \cite{tsetlin1973automaton} which investigated deterministic LA in detail. $Varshavskii$ and $Vorontsova$ \cite{varshavskii1963behavior} introduced the stochastic variable structure versions of LA. Since then LA has been extensively researched to develop various kinds of algorithms based on deterministic LA [6] and stochastic LA [7]. A comprehensive overview of these researches has been summarized by $Thathachar$ \cite{thathachar2002varieties}.

In general, the rate of convergence is one of the vital considerations of learning algorithms. Therefore, $Thathachar$ and $Sastry$ designed a new class of learning automata, called estimator algorithms \cite{thathachar1985new}\cite{thathachar1986estimator}. The estimator algorithms have faster rate of convergence than all previous ones. These algorithms, not only maintain and update the probabilities vector of actions like the previously, but also keep estimating the reward probabilities for each action with using a reward-estimator vector to update the action probability vector. In this strategy, even when an action is rewarded, it is possible that the probability of choosing another action is increased \cite{papadimitriou2004new}. Compared with the traditional learning algorithms, the estimator algorithms have been demonstrated to be more efficient. However, the performance of early estimator algorithms is strictly dependent on the reliability of the estimator’s contents and an unreliable estimator may cause a significant decrease of the accuracy and the speed of convergence \cite{papadimitriou2004new}. In this situation, $Papadimitriou$, $Sklira$ and $Pomportsis$ \cite{papadimitriou2004new} designed a stochastic estimator reward-inaction learning automaton ($SE_{RI}$) which is based on the use of a stochastic estimator. As its much faster speed of convergence and much higher accuracy in choosing the correct action than other estimator algorithms, $SE_{RI}$ is is widely accepted as the most classic LA model by now.

Due to the superiority of the estimator algorithms, there are many novel estimators \cite{ge2015novel}\cite{jiang2011new}\cite{jiang2016new} are proposed in recent years. In 2005, Hao Ge \cite{ge2015novel} proposed a deterministic estimator based LA (Discretized Generalized Confidence Pursuit Algorithm, $DGCPA$) of which the estimate of each action is the upper bound of a confidence interval and extended the algorithm to stochastic estimator schemes. The improved stochastic estimator based LA $DGCPA^{*}$ is the current fastest LA model. Although the family of estimator learning automata has achieved great improvements in the field of LA, there are still some drawbacks.

Because of the fundamental defect, the value of an estimator could not always be strictly unmistakable. Especially in the initial stage of the learning process of LA, the estimator may perform poorly on estimating the reward probabilities of each action. In this situation, a lot of reward would be added to the probabilities of non-optimal actions. Thus, a large number of extra iterations are needed to compensate for these wrong reward.

In this paper, in order to overcome the drawbacks of estimator algorithms, a novel method based on a double competitive strategy to update the action probability vector is introduced. The proposed Double Competitive Algorithm($DCA$) learning automata use a stochastic estimator as same as $SE_{RI}$. The first competitive strategy of $DCA$ is that only the action which has the highest current stochastic estimate of reward probability gets the opportunity to increase its probability. And its second strategy is that whenever the ‘optimal’ action which has the highest current stochastic estimate of reward probability changes, the probability of new ‘optimal’ action gets a huge increase while the probability of original ‘optimal’ action decreases a lot. Accordingly, the wrong rewards could be corrected instantly. Consequently, the $DCA$ learning automata converge rapidly and accurately.

The key contributions of this paper are summarized as follows.

-	We propose a new algorithm, referred to as Double Competitive Algorithm ($DCA$) and prove that the proposed scheme is $\epsilon-optimal$ in all random stationary environments.

-	The proposed $DCA$ is compared with the most classic LA $SE_{RI}$ and the fastest LA $DGCPA^{*}$ in various stationary P-model random environments. The results indicate that the proposed $DCA$ is more efficient.

The paper is organized as follows. In section 2, we introduce the general idea of LA and the estimator algorithms. The $DCA$ scheme is presented in section3. In section 4, we prove that the proposed scheme is $\epsilon-optimal$. Extensive simulation results are presented to describe the superiority of the proposed $DCA$ model over the most classic LA $SE_{RI}$ and the fastest LA $DGCPA^{*}$ in Section 5. We conclude the paper in the last section.

\section{Learning Automata and Estimator Algorithms}
\label{sec:1}
\subsection{LA and stochastic environment}
A LA is defined by a quintuple $ < A,B,Q,F( \cdot , \cdot ),G( \cdot ) > $, where:

$\bullet$ $A = \{ {\alpha _1},{\alpha _2}, \cdots ,{\alpha _r}\} $ is the set of outputs or actions, and ${\alpha _t}$ is the action chosen by the automata at any time instant $t$.

$\bullet$ $B = \{ {\beta _1},{\beta _2}, \cdots ,{\beta _m}\} $ is the set of inputs to the automata, and ${\beta _t}$ is the input at any time instant $t$. The set $t$ could be finite or infinite. In this paper, we consider the case when $B = \{0,1\} $, where $\beta = 0$ represents the events that the LA has been penalized, and $\beta = 1$ represents the events that the LA has been rewarded.

$\bullet$ $Q = \{ {q_1},{q_2}, \cdots ,{q_s}\} $ is the set of finite states, and ${q_t}$ is the state of the automata at any time instant $t$.

$\bullet$ $F( \cdot , \cdot ):Q \times B \to Q$ is a mapping in terms of the state and input at any time instant $t$, such that, $q(t + 1) = F(q(t),\beta (t))$.

$\bullet$ $G( \cdot )$ is a mapping $G:Q \to A$, and is called the output function which determines the output of the automata depending on the state ${q_t}$, such that, $\alpha (t) = G(q(t))$.

The random environment interacted with LA is defined as $ < A,B,C > $, where $A$ and $B$ has been defined above. $C = \{ {c_1},{c_2}, \cdots ,{c_r}\} $ is the set of reward probability, and ${c_i}$ corresponds to an input action ${\alpha _t}$.

\subsection{Estimator Algorithms}
For the purpose of improving the convergence rate of LA, $Thathachar$ and $Sastry$ designed a new-class of algorithms, called estimator algorithms [9][10]. These algorithms keep running estimates for each action using a reward-estimate vector and then use the estimate to update probabilities. According to the contents of estimators, the estimator algorithms could be divided into two classes, deterministic estimator algorithm and stochastic estimator algorithm.

The class of deterministic estimator is the majority of estimator algorithms, such as $D{P_{ri}}$ \cite{oommen1990discretized} and $DGPA$ \cite{agache2002generalized}. In these algorithms, the deterministic estimate vector $D'(t) = [d_i'(t), \cdots ,d_r'(t)]$ can be computed using the following formula which yields the maximum-likelihood estimate \cite{sastry1985systems}\cite{thathachar1979discretized}
\begin{equation}
d_i'(t) = \frac{{{W_i}(t)}}{{{Z_i}(t)}},\forall i = 1,2, \cdots ,r
\end{equation}
Where $W_i(t)$ is the number of times the action $\alpha_i$ has been rewarded until the current time $t$, and $Z_i(t)$ is the number of times the action $\alpha_i$ has been selected until the current time $t$.

$Papadimitrtriou$ \cite{vasilakos1995new} introduced a new type of estimator called “stochastic estimator”. The non-stationary environments indicate that the reward probability $c_i$ will vary with time instant $t$ which means the optimal action may change from time to time. In \cite{vasilakos1995new}, the author added an zero mean normally distributed random number to each action’s estimate probability. $Papadimitrtriou$ also extended the use of stochastic estimator to stationary environment \cite{papadimitriou2004new}. The implementation of the stochastic estimator in \cite{papadimitriou2004new} is to impose a random perturbation to the deterministic estimate, such that
\begin{equation}
{u_i}(t) = d_i'(t) + R_i^t
\end{equation}
where $u_i(t)$ is the stochastic estimate of reward probability of action $\alpha_i$ at time $t$, $d_i'(t)$ is the deterministic estimate of reward probability of action $\alpha_i$ at time $t$, and $R_i^t$ is a random number which is uniformly distributed in an interval. The length of the interval depends on a design parameter $\gamma$ and the number of times that action $\alpha_i$ has been selected up to time instant $t$.

%Text with citations \cite{RefB} and \cite{RefJ}.
\section{Double Competitive Algorithm}
\label{sec:2}
It is clear that, in estimator learning automata fields, the most important part is to estimate the reward probabilities of each possible action accurately. However, Because of the fundamental defect, the value of an estimator could not always be strictly unmistakable. Especially in the initial stage of the learning process of LA, the estimator may perform poorly on estimating the reward probabilities of each action. Thus, a lot of rewards would be added to the probabilities of non-optimal actions. As a result, a large number of extra iterations are needed to compensate for these wrong rewards.

The proposed double competitive algorithm ($DCA$) is a learning automaton which updates the action probability with a double competitive strategy to update the action probability. The first competitive strategy of $DCA$ is that only the action which has the highest current stochastic estimate of the reward probability gets the opportunity to increase its probability. And the second competitive strategy is that whenever the ‘optimal’ action which has the highest current stochastic estimate of reward probability changes, the probability of new ‘optimal’ action gets a huge increase while the probability of the original ‘optimal’ action decreases a lot. With the unique two competitive strategies, the wrong rewards could be corrected instantly. Clearly, the ‘optimal’ action would be constantly changing as the estimator is not reliable enough in the early stages of learning, resulting in the probability of each action fluctuates continually. But eventually, when the estimator is fully reliable, as the action which has the highest current stochastic estimate of reward probability tends to be invariable, the LA will converge rapidly.

Besides, since the dramatic changes of the probabilities of any possible action during the learning process, the actions whose probabilities used to be relatively small get more opportunities to be selected. Then their deterministic estimates would be further updated. Therefore, during the learning process, the estimate of each non-optimal action gets more opportunities to be updated. According to the Law of Large Numbers, the precision of the stochastic estimator would be higher. So the stochastic estimator in $DCA$ scheme would be more reliable than that in $SE_{RI}$ scheme.

The procedure of $DCA$ is briefly introduced below.

\textbf{The $DCA$ scheme}

\textbf{Algorithm $DCA$}

\textbf{Parameters}

$n,\gamma$ resolution parameters

$\mu $ attenuation factor

${W_i}(t)$ the number of events that $i$th action has been rewarded up to time instant $t$, for $1 \le i \le r$

${Z_i}(t)$ the number of events that the $i$th action has been selected up to time instant $t$, for $1 \le i \le r$

$\Delta = 1/r/n$ smallest step size

${m_0}$ the action that has the highest stochastic estimate of reward probability at the last time $t-1$.

\textbf{Method}

\textbf{Initialize} $\mu $=0.1

\textbf{Initialize} ${p_i}(t) = 1/r$ for $1 \le i \le r$

\textbf{Initialize} ${W_i}(t)$ and ${Z_i}(t)$ by selecting each action a number of times

\textbf{Initialize} ${m_0} = $ a random integer within $[1,r]$

\textbf{Repeat}

Step 1: At time $t$, choose an action $\alpha (t) = \alpha $, according to the probability distribution $P(t)$.

Step 2: Receive a feedback $\beta (t) \in \{ 0,1\} $ from stochastic environment.

Step 3: Set $W{}_i(t) = {W_i}(t - 1) + \beta (t)$, ${Z_i}(t) = {Z_i}(t - 1) + 1$.

Step 4: Compute the deterministic estimate $d_i'(t)$, by setting $d_i'(t) = \frac{{{W_i}(t)}}{{{Z_i}(t)}}$.

Step 5: If $\beta (t) = 0$, go to Step 9.

Step 6: Compute stochastic estimates ${u_i}(t) = d_i'(t) + {R_i}(t)$, where ${R_i}(t)$ is a random number uniformly distributed within $( - \frac{\gamma }{{{Z_i}(t)}},\frac{\gamma }{{{Z_i}(t)}})$.

Step 7: Select the action ${\alpha _m}$ that has the highest stochastic estimate of reward probability, where ${\alpha _m} = \max \{ {u_i}(t)\} $.

Step 8: Update the probability vector $P(t)$ according to the following equations:
\[\begin{array}{l}
{p_i}(t) = \max \{ {p_i}(t) - \Delta \} ,\forall i \ne m;\\
{p_m}(t) = 1 - \sum\limits_{i \ne m} {{p_i}(t)} .
\end{array}\]

Step 9: Compute stochastic estimates $u_i'(t)$ in the same way with Step 6 and select the action ${\alpha _m'}$ like Step 7 where ${\alpha _m'} = \max \{ u_i'(t)\} $.

Step 10: if $m' = {m_0}$, go to Step 12.

Step 11: Update the probability vector $P(t)$ as follows
\[\begin{array}{l}
{p_{m'}}(t) = {p_{m'}}(t) + (1 - \mu )*{p_{{m_0}}}(t);\\
{p_{{m_0}}}(t) = \mu *{p_{{m_0}}}(t);\\
{m_0} = m'.
\end{array}\]

Step 12: Update the probability vector $P(t)$ at time $t+1$.
\[{p_i}(t + 1) = {p_i}(t),1 \le i \le r\]
\textbf{End Repeat} \\
\textbf{End Algorithm $DCA$} \\

Note that the double competitive strategy is reflected in the twice updating probability procedures. Step 7 and Step 8 are the implementation of the first competitive strategy, only the action   which has the highest current stochastic estimate of the reward probability gets the opportunity to increase its probability and in order to satisfy $\sum_{i = 1}^r {p_i(t) = 1} $, all the probabilities of the others decrease. The second competitive strategy is summarized in Step 10 and Step 11 where whenever the ‘optimal’ action which has the highest current stochastic estimate of reward probability changes ($m' \ne {m_0}$), the probability of the original ‘optimal’ action ${\alpha _{{m_0}}}$ is reduced by $90\%$ (determined by the attenuation factor $\mu$), and then the new ‘optimal’ action ${\alpha _{m'}}$ will get an additional reward which equals to the reduced probability of action ${\alpha _{{m_0}}}$.

\section{Proof of $\epsilon-optimality$}
\label{sec:3}
Whether the given algorithm is $\epsilon-optimality$ is an important standard in LA contexts. Thus, we will show that the proposed $DCA$ scheme is $\epsilon-optimal$ in every stationary environment.\\
\textbf{Definition 1} $\epsilon-optimality$ Given any arbitrarily small $\varepsilon > 0$ and $\delta > 0$, there exists a ${n_0} < \infty $ (that depends on $\varepsilon$ and $\delta $) and a ${t_0} < \infty $ such that for all resolution parameter $n \ge {n_0}$ and all time $t \ge {t_0}$ :$\Pr \{ |p{}_m(t) - 1| < \varepsilon \}  > 1 - \delta$.\\
To prove the $\epsilon-optimality$ of $DCA$ scheme, the following two theorems would be used.\\
\textbf{Theorem 1}:Suppose there exists an index $m$ and a time instant ${t_0} < \infty $ such that ${u_m}(t) > {u_j}(t)$ for all $j$ with $j \ne m$ and for all time $t \ge {t_0}$. Then there exists an integer $n_0$ such that for all resolution parameters $n > {n_0}$, ${p_m}(t) \to 1$ with probability one as $t \to \infty $.\\
\textbf{Proof}: Since we have supposed that ${u_m}(t) > {u_j}(t)$ for all $j$ with $j \ne m$ and for all time $t \ge t_0$, the action $\alpha_m$ which has the highest stochastic estimate of reward probability will not change, there is no difference between the proposed $DCA$ and $SE_{ri}$ scheme. The scheme has been introduced and proved in [11].\\
\textbf{Theorem 2}:For action $\alpha_i$， assume ${p_i}(0) \ne 0$, for any given constants $\delta  > 0$ and ${\rm M} > 0$, there exists ${n_0} < \infty $ and ${t_0} < \infty $ such that for all resolution parameters $n \ge {n_0}$ and all time $t \ge {t_0}$: $Pr [{\alpha _i}{\rm{\ is \  selected\ more\ than\ M\ times\ at\ time\ t]}} \ge {\rm{1 - }}\delta $.

\textbf{Proof}:Define the random variable ${\alpha _i}$ as the number of times the action ${\alpha _i}$ is chosen up to time instant $t$. And then we should prove that
\begin{equation}
\Pr \{ G_i^t > M\}  \ge 1 - \delta .
\end{equation}

Equivalent to prove that

\begin{equation}
\Pr \{ G_i^t \le M\}  \le \delta .
\end{equation}

It is clear that the events $\{ G_i^t = k\} $ and $\{ G_i^t = j\} $ are mutually exclusive for any $j \ne k$.Then [4] is equivalent to

\begin{equation}
\sum\limits_{k = 1}^M {\Pr \{ G_i^t = k\} }  \le \delta .
\end{equation}

Now, consider an extreme situation in the proposed $DCA$ learning automata. If the random initialization ${m_0} = i$ and the action ${\alpha _i}$ does not have the highest stochastic estimate of reward probability in the first iteration, then the probability of action ${\alpha _i}$ gets a ninety percent decay. And worse, the $i$th action would not get any reward in the subsequent iterations which means the stochastic estimate of reward probability of action ${\alpha _i}$ meets that ${u_i}(t) < {u_m}(t)$ at all time instant $t$. Thus, during any of the first $t$ iterations, the largest decrease for any action is $(1-\mu)*p{}_i(0) + t*\Delta $. So it is clear that:

\begin{equation}
\Pr \{ \alpha {}_i{\rm{\ is\ not\ chosen\} }} \le {\rm{(1}} - \mu*{p_i}(0) + t*\Delta ).
\end{equation}

The probability that action ${\alpha_i}$ is chosen up to $M$ times among $t$ iterations has the following upper bound.

\begin{equation}
\Pr \{ G_i^t \le M{\rm{\} }} \le \sum\limits_{k = 1}^M {C(t,k){{(1)}^k}(} {\rm{1}} - \mu*{p_i}(0) + t*\Delta {)^{t - k}}.
\end{equation}

It is clear that a sum of $M$ terms would less than $\delta $ if each element of the sum less than $\delta /M$. And when $k = m$, $C(t,m) \le {t^m}$. Thus we should prove that:

\begin{equation}
M{t^m}{{\rm{(1}} - \mu*{p_i}(0) + t*\Delta )^{t - m}} \le \delta .
\end{equation}

Observe the inequality, it is necessary to make sure that ${\rm{(1}} - \mu*{p_i}(0) + t*\Delta )$ is strictly is less than unity when $t$ increases. Thus $\Delta  > \frac{{\mu*{p_i}(0)}}{t}$ with $\Delta  = \frac{1}{{rn}}$, such that $\Delta  > \frac{{\mu *{p_i}(0)}}{t}$. Let,

\begin{equation}
n = \frac{{2t}}{{\mu rp{}_i(0)}}.
\end{equation}

Now, we should prove that

\begin{equation}
\Pr \{ G_i^t \le M\}  \le M{t^m}{\Psi ^{t - m}}.
\end{equation}

Where

\begin{equation}
\Psi  = 1 - \frac{\mu }{2}{{\rm{p}}_i}{\rm{(0)  and  }}0 < \Psi  < 1.
\end{equation}

Then we calculate that

\begin{equation}
\mathop {\lim }\limits_{t \to \infty } M{t^m}{\Psi ^{t - m}} = M{\lim _{t \to  \propto }}\frac{{{t^m}}}{{{{(1/\Psi )}^{t - m}}}}.
\end{equation}

Using l’Hopital's rule $m$ times, we could get the following equation:

\begin{equation}
M\mathop {\lim }\limits_{t \to \infty } \frac{{{t^m}}}{{{{(1/\Psi )}^{t - m}}}} = M\mathop {\lim }\limits_{t \to \infty } \frac{{m!}}{{(\ln {{(1/\Psi )}^m}{{(1/\Psi )}^{t - m)}}}} = 0.
\end{equation}

Thus, $M\mathop {\lim }\limits_{t \to \infty } \frac{{{t^m}}}{{{{(1/\Psi )}^{t - m}}}}$ has a limit of zero as $t$ tends towards infinity with $n = \frac{{2t}}{{\mu rp{}_i(0)}}$. In this case, for every action ${\alpha _i}$, there is a ${t_i}$, and for all $t > t(i)$, $M{t^m}{{\rm{(1}} - \mu*{p_i}(0) + t*\Delta )^{t - m}}$ is less than $\delta$. And, it’s clear that (8) is monotonically decreasing as $n$ increases. Let $n = \frac{{2t}}{{\mu rp{}_i(0)}}$. Hence, (8) is satisfied for all $n > {n_i}$ and $t > {t_i}$.Furthermore, for any $t > {t_i}$, we have

\begin{equation}
\Pr \{ G_i^t \ge M\}  \ge \Pr \{ G_i^{{t_i}} \ge M\} .
\end{equation}

Thus, we could get that

\begin{equation}
G_i^{{t_i}} \ge M \Rightarrow G_i^t \ge M.
\end{equation}

Hence, for any action ${\alpha _i}$,

\begin{equation}
\Pr \{ G_i^t \le M\}  \le \delta {\rm{ whenever }}t > {t_i}{\rm{ and }}n > {n_i}.
\end{equation}

Now, we could repeat this argument for all the actions. Define ${n_0}$ and ${t_0}$ as follows:

\[\begin{array}{l}
{n_0} = {\max _{1 \le i \le r}}\{ {n_i}\} ;\\
{t_0} = {\max _{1 \le i \le r}}\{ {t_i}\} .
\end{array}\]

Thus, for each action, $\Pr \{ G_i^t < M\}  < \delta$ is satisfied for all $t > {t_0}$ and $n > {n_0}$, and the theorem is proved.

Now we are ready to prove that $DCA$ scheme is $\epsilon-optimal$. According to the Definition 1, we should prove the following theorem.

\textbf{Theorem 3}:The $DCA$ is $\epsilon-optimal$ in every random environment. Given any $\varepsilon > 0$ and $\delta > 0$, there exists a ${n_0} < \infty $(that depends on $\varepsilon $ and $\delta $)and ${t_0} < \infty $ such that for all $n > {n_0}$ and $t > {t_0}$: $\Pr \{ |{P_m}(t) - 1| < \varepsilon \}  > 1 - \delta $.

\textbf{Proof}: The only difference between the proposed $DCA$ scheme and $SE_{ri}$ scheme is the method to update the probabilities. Since we have shown that the theorem 1 and theorem 2 work well in $DCA$, we can prove the $\epsilon-optimality$ of $DCA$ in the same method with $SE_{ri}$ which has been introduced in detail in \cite{papadimitriou2004new}.\\
\section{Simulation results}
\label{sec:4}

In the following, the proposed $DCA$ scheme is compared with the most classic LA $SE_{RI}$ and $DGCPA^{*}$ which is considered as the current fastest LA. All of the schemes have been proved to be  $\epsilon-optimal$.

Within the context of LA, the speed of convergence is compared by the iterations needed to converge under the five benchmark environments given in \cite{papadimitriou2004new}. The actions’ reward probabilities for each environment are as follows:

\begin{itemize}
  \item $E_1$: D=\{0.65,0.50,0.45,0.40,0.35,0.30,0.25,0.20,0.15,0.10\}.
  \item $E_2$: D=\{0.60,0.50,0.45,0.40,0.35,0.30,0.25,0.20,0.15,0.10\}.
  \item $E_3$: D=\{0.55,0.50,0.45,0.40,0.35,0.30,0.25,0.20,0.15,0.10\}.
  \item $E_4$: D=\{0.70,0.50,0.30,0.20,0.40,0.50,0.40,0.30,0.50,0.20\}.
  \item $E_5$: D=\{0.10,0.45,0.84,0.76,0.20,0.40,0.60,0.70,0.50,0.30\}.
\end{itemize}

In all the simulations performed, we have the same setting as \cite{papadimitriou2004new}. The computation of an algorithm is considered to have converged if the probability of choosing an action is greater than or equal to a threshold $T$($0 < T < 1$). The automaton is considered to have converged correctly when it converges to the action that has the highest reward probability.

Before comparing the performance of different learning automata, a large number of evaluation tests were carried out to determine the ‘best’ parameters for each scheme. The values of ‘best’ parameters are considered to be the best if they yield the fastest convergence and the automaton converges to the correct action in a sequence of   experiments. The values of $DCA$ and $SE_{RI}$ are taken to the same as those used in \cite{papadimitriou2004new}. Hence, $T = 0.999$ and $NE = 750$. As long as we have determined the ‘best’ parameters, each algorithm was executed 250,000 times for each environment by using the ‘best’ parameters. Before the simulation, to initialize the estimator vector, all the actions were sampled 10 times, and these extra 100 iterations are included in the iteration counts.

\begin{figure*}
% Use the relevant command to insert your figure file.
% For example, with the graphicx package use
  \includegraphics[width=12cm]{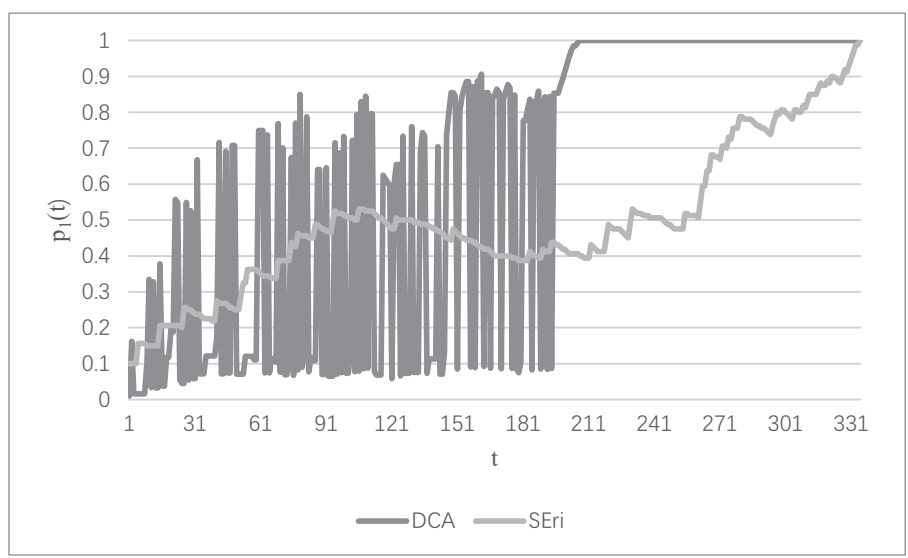}
% figure caption is below the figure
\caption{${p_1}(t)$ versus $t$(the extra 100 iterations used to initialize the estimator vector is not included) characteristics of $DCA$ and $SE_{ri}$ when operating in environment $E_1$. For both schemes, the ‘best’ learning parameters are used.}
\label{fig:2}       % Give a unique label
\end{figure*}

Before comparing the overall simulation results, a single ordinary experiment would be executed to show the difference between $DCA$ and $SE_{ri}$ during the convergence process. The curves that represent the probability of the optimal action as a function of time are presented in Fig \ref{fig:2}.

\begin{figure*}
% Use the relevant command to insert your figure file.
% For example, with the graphicx package use
  \includegraphics[width=12cm]{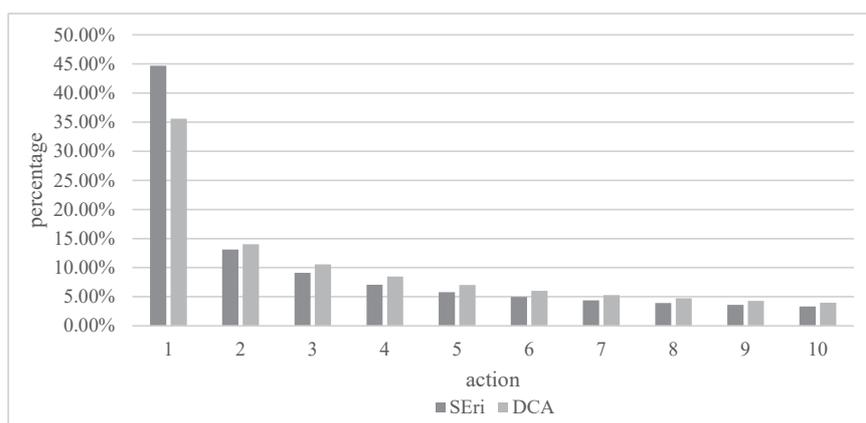}
% figure caption is below the figure
\caption{the percentage of the number of each action has been selected to the total number required for convergence of $DCA$ and $SE_{ri}$ in environment $E_1$ , when using the ‘best’ learning parameters(250,000 experiments were performed for each scheme).}
\label{fig:3}       % Give a unique label
\end{figure*}

The results presented in Fig \ref{fig:2} indicate that the probability of the optimal action changes dramatically in the initial stage of the $DCA$ learning process as we have explained earlier. With the number of iterations increasing, the stochastic estimator becomes more and more reliable. When the estimator is sufficiently reliable, the learning automaton converges rapidly. On the other hand, during the convergence process of $SE_{RI}$ scheme, once the probability of the optimal action decreases, a lot of extra iterations are needed to compensate for the lost probability.\\

Besides, as presented in Fig \ref{fig:3}, the non-optimal actions in $DCA$ scheme have more chances to be selected than in $SE_{RI}$ scheme. Thus, during the learning process, the estimate of each action gets more opportunities to be updated and the precision of the stochastic estimator would be higher. So the time $t$ when the stochastic estimator is reliable enough in $DCA$ scheme would be earlier than that in $SE_{ri}$ scheme.

Thus, with the benefits that have been explained above, the overall simulation results are presented as follows.

\begin{table}
% table caption is above the table
\caption{Accuracy (number of correct convergences/number of experiments) of $DCA$ and $SE_{RI}$ in environment $E_1$ to $E_5$, when using the ‘best’ learning parameters(250,000 experiments were performed for each scheme in each environment)}
\label{tab:1}       % Give a unique label
% For LaTeX tables use
\begin{tabular}{llllll}
\hline\noalign{\smallskip}
  & $E_1$ & $E_2$ & $E_3$ & $E_4$ & $E_5$  \\
\noalign{\smallskip}\hline\noalign{\smallskip}
$DCA$ & 0.998 & 0.997 & 0.996 & 0.999 & 0.998\\
$SE_{RI}$ & 0.997 & 0.996 & 0.995 & 0.998 & 0.997  \\
$DGCPA^{*}$ & 0.997 & 0.996 & 0.995 & 0.998 & 0.997  \\
\noalign{\smallskip}\hline
\end{tabular}
\end{table}

\begin{table}
% table caption is above the table
\caption{Comparison of the average number of iterations required for convergence of $DCA$ and $SE_{ri}$ in environment $E_1$ to $E_5$, when using the ‘best’ learning parameters(250,000 experiments were performed for each scheme in each environment)}
\label{tab:2}       % Give a unique label
% For LaTeX tables use
\begin{tabular}{lllllll}
\hline\noalign{\smallskip}
Environment & $DCA$ &   & $SE_{RI}$ &   & $DGCPA$ &\\
\cline{2-3}\cline{4-5}\cline{6-7}
  & Parameter & Iterations & Parameter & Iterations & Parameter & Iterations\\
\cline{1-7}
$E_1$ & $n=13 ,\gamma=6 $ & 377  & $n=16 ,\gamma=8 $ & 426 & $n=3 ,\gamma=5 $ & 351  \\
$E_2$ & $n=23 ,\gamma=8 $ &  664 & $n=32 ,\gamma=12$ &  834 & $n=6 ,\gamma=9 $ & 678 \\
$E_3$ & $n=43 ,\gamma=16 $ &  2134 & $n=105 ,\gamma=25 $ & 2540 & $n=19 ,\gamma=20 $ & 2032  \\
$E_4$ & $n=12 ,\gamma=5 $ & 299  & $n=13 ,\gamma=6 $ & 325 & $n=2 ,\gamma=4 $ & 298  \\
$E_5$ & $n=40 ,\gamma=7 $ &  633 & $n=33 ,\gamma=12 $ & 729 & $n=5 ,\gamma=7 $ & 598  \\
\cline{1-7}
\end{tabular}
\end{table}

\begin{table}
% table caption is above the table
\caption{Comparison of the average number of iterations required for convergence achieving the same accuracy as $SE_{RI}$ shown in Table 1 in environment $E_1$ to $E_5$(250,000 experiments were performed for each scheme in each environment)}
\label{tab:3}       % Give a unique label
% For LaTeX tables use
\begin{tabular}{llllll}
\hline\noalign{\smallskip}
Environment & $DCA$ &   & $SE_{RI}$ &   & Improvement\\
\cline{2-3}\cline{4-5}
  & Parameter & Iterations & Parameter & Iterations\\
\cline{1-6}
$E_1$ & $n=10 ,\gamma=6 $ & 338 & $n=16 ,\gamma=8 $ & 426 & 20.66\%\\
$E_2$ & $n=18 ,\gamma=8 $ &  633 & $n=32 ,\gamma=12 $ &  834 & 24.10\%\\
$E_3$ & $n=30 ,\gamma=16 $ & 1990 & $n=105 ,\gamma=25 $ &  2540 & 21.65\%\\
$E_4$ & $n=9 ,\gamma=5 $ & 282 & $n=13 ,\gamma=6 $ & 325 & 13.23\%\\
$E_5$ & $n= 28,\gamma=7 $ & 582 & $n=33 ,\gamma=12 $ & 729 & 20.16\%\\
\cline{1-6}
\end{tabular}
\end{table}

\begin{table}
% table caption is above the table
\caption{Comparison of the average number of iterations the average time required for convergence achieving the same accuracy as $SE_{RI}$ shown in Table 1 in environment $E_1$ to $E_5$(250,000 experiments were performed for each scheme in each environment)}
\label{tab:4}       % Give a unique label
% For LaTeX tables use
\begin{tabular}{lllllll}
\hline\noalign{\smallskip}
Environment & $DCA$ &   &   & $DGCPA^{*}$ \\
\cline{2-4}\cline{5-7}
  & Parameter & Iterations & Time(ms) & Parameter & Iterations & Time(ms)\\
\cline{1-7}
$E_1$ & $n=10 ,\gamma=6 $ & 338 & 0.162 & $n=16 ,\gamma=8 $ & 426 & 3.423\\
$E_2$ & $n=18 ,\gamma=8 $ &  633 & 0.339 & $n=32 ,\gamma=12 $ &  834 & 7.417\\
$E_3$ & $n=30 ,\gamma=16 $ & 1990 & 1.167 & $n=105 ,\gamma=25 $ &  2540 & 26.577\\
$E_4$ & $n=9 ,\gamma=5 $ & 282 & 0.126 & $n=13 ,\gamma=6 $ & 325 & 2.744\\
$E_5$ & $n= 28,\gamma=7 $ & 582 & 0.351 & $n=33 ,\gamma=12 $ & 729 & 9.252\\
\cline{1-7}
\end{tabular}
\end{table}

The accuracies (number of correct convergences/number of experiments) of $DCA$, $SE_{RI}$ and $DGCPA^{*}$ in environment $E_1$ to $E_5$ when using the ‘best’ learning parameters are presented in Table \ref{tab:1}. The results show that $DCA$ always has a better accuracy than the other two algorithms. The average numbers of iterations required for convergence are summarized in Table \ref{tab:2}, which demonstrate that the $DCA$ scheme converges with a faster speed than $SE_{RI}$ and with a little lower speed than the current fastest LA $DGCPA^{*}$ with higher accuracy. In order to ensure that the performance comparison between $DCA$, $SE_{RI}$ and $DGCPA^{*}$ is fair, let us verify the number of iterations required to achieve the same accuracy, a series of experiments have been carried out. The results are shown in Table \ref{tab:3} and Table \ref{tab:4}.

On the one hand, compared with the most classic LA model $SE_{RI}$, the proposed scheme $DCA$ achieves a great improvement in the speed of convergence in all benchmark environments. For example, in environment $E_2$, The $DCA$ converges in 633 iterations, while the $SE_{RI}$ requires 834 iterations. Thus, an improvement of $24.10\%$ in comparison with $SE_{RI}$ is obtained.

On the other hand, as indicated in Table \ref{tab:4}, the current fastest LA model $DGCPA^{*}$ performs less competitively than the proposed $DCA$ scheme. The superiority of $DCA$ is not only reflected in the fewer number of iterations for convergence, but also established in the time efficiency. Because of the complexity of $DGCPA^{*}$ model when computing the confidence interval, the time required for convergence increases rapidly. Thus, the superiority of the proposed $DCA$ scheme is clear.

In summary, the $DCA$ scheme using a double competitive strategy is more efficient than $SE_{RI}$ and $DGCPA^{*}$. It overcomes the drawbacks of estimator algorithms and provides a novel idea to make breakthroughs in LA fields.

\section{Conclusions}
\label{sec:5}
In this paper, a novel P-model absorbing learning automaton is introduced. With the use of double competitive strategy, the proposed $DCA$ scheme overcomes the drawbacks of existing estimator algorithms. The benefits of proposed scheme are analysed and it is proved to be $\epsilon-optimal$ in every stationary random environment. Extensive simulations have been performed in five benchmark environments, and the results indicate that the proposed $DCA$ scheme converges faster and performs more efficiently than the most classic LA $SE_{RI}$ and the current fastest LA $DGCPA^{*}$. Since the reliability of an estimator is the key to guarantee the convergence of LA, the future work will focus on studying how to make the estimator being reliable enough as soon as possible.

\begin{acknowledgements}
This research work is funded by the National Key Research and Development Project of China (2016YFB0801003), Science and Technology Project of State Grid Corporation of China ( SGCC)，Key Laboratory for Shanghai Integrated Information Security Management Technology Research.
\end{acknowledgements}

 %BibTeX users please use one of
%\bibliographystyle{spbasic}      % basic style, author-year citations
\bibliographystyle{spmpsci}      % mathematics and physical sciences
\bibliography{template}   % name your BibTeX data base

\begin{thebibliography}{10}
\providecommand{\url}[1]{{#1}}
\providecommand{\urlprefix}{URL }
\expandafter\ifx\csname urlstyle\endcsname\relax
  \providecommand{\doi}[1]{DOI~\discretionary{}{}{}#1}\else
  \providecommand{\doi}{DOI~\discretionary{}{}{}\begingroup
  \urlstyle{rm}\Url}\fi

\bibitem{agache2002generalized}
Agache, M., Oommen, B.J.: Generalized pursuit learning schemes: new families of
  continuous and discretized learning automata.
\newblock IEEE Transactions on Systems, Man, and Cybernetics, Part B
  (Cybernetics) \textbf{32}(6), 738--749 (2002)

\bibitem{esnaashari2010data}
Esnaashari, M., Meybodi, M.R.: Data aggregation in sensor networks using
  learning automata.
\newblock Wireless Networks \textbf{16}(3), 687--699 (2010)

\bibitem{ge2015novel}
Ge, H., Jiang, W., Li, S., Li, J., Wang, Y., Jing, Y.: A novel estimator based
  learning automata algorithm.
\newblock Applied Intelligence \textbf{42}(2), 262--275 (2015)

\bibitem{jiang2011new}
Jiang, W.: A new class of $\varepsilon$-optimal learning automata.
\newblock In: International Conference on Intelligent Computing, pp. 116--121.
  Springer (2011)

\bibitem{jiang2016new}
Jiang, W., Li, B., Li, S., Tang, Y., Chen, C.L.P.: A new prospective for
  learning automata: A machine learning approach.
\newblock Neurocomputing \textbf{188}, 319--325 (2016)

\bibitem{jiang2014new}
Jiang, W., Zhao, C.L., Li, S.H., Chen, L.: A new learning automata based
  approach for online tracking of event patterns.
\newblock Neurocomputing \textbf{137}, 205--211 (2014)

\bibitem{narendra2012learning}
Narendra, K.S., Thathachar, M.A.: Learning automata  (2012)

\bibitem{nicopolitidis2002using}
Nicopolitidis, P., Papadimitriou, G.I., Pomportsis, A.S.: Using learning
  automata for adaptive push-based data broadcasting in asymmetric wireless
  environments.
\newblock IEEE Transactions on vehicular technology \textbf{51}(6), 1652--1660
  (2002)

\bibitem{oommen1996graph}
Oommen, B.J., Croix, E.d.S.: Graph partitioning using learning automata.
\newblock IEEE Transactions on Computers \textbf{45}(2), 195--208 (1996)

\bibitem{oommen1990discretized}
Oommen, B.J., Lanct{\^o}t, J.K.: Discretized pursuit learning automata.
\newblock IEEE Transactions on systems, man, and cybernetics \textbf{20}(4),
  931--938 (1990)

\bibitem{papadimitriou2004new}
Papadimitriou, G.I., Sklira, M., Pomportsis, A.S.: A new class of
  \&epsi;-optimal learning automata.
\newblock IEEE Transactions on Systems, Man, and Cybernetics, Part B
  (Cybernetics) \textbf{34}(1), 246--254 (2004)

\bibitem{sastry1985systems}
Sastry, P.: Systems of learning automata: Estimator algorithms applications.
\newblock Ph.D. thesis, Ph. D. Thesis, Dept of Electrical Engineering, Indian
  Institute of Science, Bangalore, India (1985)

\bibitem{thathachar1979discretized}
Thathachar, M., Oommen, B.: Discretized reward-inaction learning automata.
\newblock J. Cybern. Inf. Sci \textbf{2}(1), 24--29 (1979)

\bibitem{thathachar1985new}
Thathachar, M., Sastry, P.S.: A new approach to the design of reinforcement
  schemes for learning automata.
\newblock IEEE Transactions on Systems, Man, and Cybernetics (1), 168--175
  (1985)

\bibitem{thathachar1986estimator}
Thathachar, M.A., Sastry, P.S.: Estimator algorithms for learning automata
  (1986)

\bibitem{thathachar2002varieties}
Thathachar, M.A., Sastry, P.S.: Varieties of learning automata: an overview.
\newblock IEEE Transactions on Systems, Man, and Cybernetics, Part B
  (Cybernetics) \textbf{32}(6), 711--722 (2002)

\bibitem{tsetlin1973automaton}
TSetlin, M., et~al.: Automaton theory and modeling of biological systems
  (1973)

\bibitem{varshavskii1963behavior}
Varshavskii, V., Vorontsova, I.: On the behavior of stochastic automata with a
  variable structure.
\newblock Avtomatika i Telemekhanika \textbf{24}(3), 353--360 (1963)

\bibitem{vasilakos1995new}
Vasilakos, A.V., Papadimitriou, G.I.: A new approach to the design of
  reinforcement schemes for learning automata: Stochastic estimator learning
  algorithm.
\newblock Neurocomputing \textbf{7}(3), 275--297 (1995)

\bibitem{wang2014learning}
Wang, Y., Jiang, W., Ma, Y., Ge, H., Jing, Y.: Learning automata based
  cooperative student-team in tutorial-like system.
\newblock In: International Conference on Intelligent Computing, pp. 154--161.
  Springer (2014)

\bibitem{zhao2015cellular}
Zhao, Y., Jiang, W., Li, S., Ma, Y., Su, G., Lin, X.: A cellular learning
  automata based algorithm for detecting community structure in complex
  networks.
\newblock Neurocomputing \textbf{151}, 1216--1226 (2015)

\end{thebibliography}

% Non-BibTeX users please use
%\begin{thebibliography}{}
%%
%% and use \bibitem to create references. Consult the Instructions
%% for authors for reference list style.
%%
%%\bibitem{RefJ}
%%% Format for Journal Reference
%%Narendra, Kumpati S., and Mandayam AL Thathachar. "Learning automata." (2012).
%%% Format for books
%%\bibitem{RefB}
%%Barto, Andrew G., and P. Anandan. "Pattern-recognizing stochastic learning automata." IEEE Transactions on Systems, Man, and Cybernetics 3 (1985): 360-375.
%% etc
%\end{thebibliography}

\end{document}